\documentclass[runningheads]{llncs}
\usepackage{graphicx}
\usepackage{multirow}
\usepackage{subcaption}
\usepackage{tikz}
\usepackage{amsmath}
\usepackage{amssymb}

\def\scalevaltab{0.75}

\begin{document}

\title{Unleashing the Power of Dynamic Mode Decomposition and Deep Learning for Rainfall Prediction in North-East India. } 
\titlerunning{Rainfall prediction}
%
\author{Paleti Nikhil Chowdary\inst{1}\orcidID{0009-0002-2300-0997} \and Sathvika P\inst{1}\orcidID{0009-0001-8437-010X} \and
Pranav U\inst{1}\orcidID{0009-0007-6643-3326} \and Rohan S\inst{1}\orcidID{0009-0005-7440-2023}  \and Sowmya V\inst{1}\orcidID{0000-0003-3745-6944}  \and Gopalakrishnan E A\inst{1,2}\orcidID{0000-0003-3689-6083} \and Dhanya M\inst{3}\orcidID{0000-0002-6958-2825} }
\authorrunning{Nikhil Chowdary .P et al.}

\institute{Amrita School of Artificial Intelligence, Coimbatore, Amrita Vishwa Vidyapeetham, India.
\\ \and
Department of Computer Science and Engineering, Amrita School of Computing, Amrita Vishwa Vidyapeetham, Bangalore, India. \\ \and
Center for Wireless Networks and Applications (WNA), Amrita Vishwa Vidyapeetham, Amritapuri, India.}

\maketitle 
\begin{abstract}

Accurate rainfall forecasting is crucial for effective disaster preparedness and mitigation in the North-East region of India, which is prone to extreme weather events such as floods and landslides. In this study, we investigated the use of two data-driven methods, Dynamic Mode Decomposition (DMD) and Long Short-Term Memory (LSTM), for rainfall forecasting using daily rainfall data collected from India Meteorological Department in northeast region over a period of 118 years. We conducted a comparative analysis of these methods to determine their relative effectiveness in predicting rainfall patterns. Using historical rainfall data from multiple weather stations, we trained and validated our models to forecast future rainfall patterns. Our results indicate that both DMD and LSTM are effective in forecasting rainfall, with LSTM outperforming DMD in terms of accuracy, revealing that LSTM has the ability to capture complex nonlinear relationships in the data, making it a powerful tool for rainfall forecasting. Our findings suggest that data-driven methods such as DMD and deep learning approaches like LSTM can significantly improve rainfall forecasting accuracy in the North-East region of India, helping to mitigate the impact of extreme weather events and enhance the region's resilience to climate change.


\end{abstract}
\section{Introduction}


Rainfall is one of the most important climatic variables that affects various aspects of human life and natural ecosystems. Accurate rainfall forecasting is crucial for effective disaster preparedness and mitigation. The North-East region of India, commonly known as the ``Seven Sisters'' is a topographically diverse region with a unique mix of flora and fauna. However, it is also one of the most vulnerable regions in the world, with a high incidence of natural disasters, particularly floods and landslides. In this region, accurate rainfall forecasting is essential for effective disaster preparedness and mitigation, particularly in the face of increasing occurrences of extreme weather events.


The North-East region receives the highest annual rainfall in India, with its hilly terrain increasing its susceptibility to landslides and flash floods. The region also experiences cyclones and thunderstorms, which have the potential to cause extensive damage to infrastructure and agriculture. Climate change has exacerbated the frequency and intensity of these weather events, leading to prolonged dry spells and erratic rainfall patterns, further exacerbating the challenges faced by the region.

Rainfall prediction methods in the past relied on empirical relationships between atmospheric variables (temperature, humidity, wind, and pressure) using statistical \cite{ref15} and dynamical approaches involving computer simulations \cite{ref16}. However, their effectiveness was limited due to complex interactions and uncertainty. Recently, there is increasing interest in using machine learning (ML) and data-driven techniques to enhance rainfall predictions \cite{ref17}.

The present work aims to build data driven models to predict rainfall in mm using monthly average rainfall data. The work involved implementing Deep Learning and Dynamic Mode Decomposition techniques and performing various experiments to determine the hyperparameters that yield the best results. 


The dataset from India Meteorological Department (IMD) is considered in this study, The dataset is processed to construct average monthly data and a DMD model is build using it. Few key locations are considered from the chosen region and a sliding-window based Deep Learning(DL) model is built using it. The models are then evaluated on RMSE and MAE metrics. 

Our results show that both DMD techniques and LSTM techniques are capable to capture the patterns in the data allowing the models to forecast rainfall effectively. Our DMD method got MSE values ranging
 from 150.44 mm to 263.34 mm and MAE values ranging from 91.34 mm to 154.61 mm while the DL approach on average got a normalized MAE value of 0.35 and a normalized RMSE value of 0.534.

The paper is structured as follows: section 2 provides an overview of the existing literature, section 3 outlines the proposed methodology, section 4 presents the results and corresponding discussion, and section 5 provides the conclusion.

\section{Literature review}

In recent years, several deep learning-based approaches have been proposed for rainfall forecasting. In the paper “a deep convolutional neural network with bi-directional long short-term memory model for short-term rainfall prediction” Convolutional Neural Networks (CNNs) combined with Long Short-Term Memory (LSTM) networks have been used to create a new network called Tiny-RainNet which can be used for rainfall forecasting from radar images (Zhang et al)\cite{ref11}.However, this approach only predicts rainfall only one or two hours into the future.

Some recent studies have proposed incorporating location information into deep learning-based rainfall forecasting models. For example, (Men et al)\cite{ref12} in the paper “Spatio-temporal Analysis of Precipitation and Temperature: A Case Study Over the Beijing–Tianjin–Hebei Region, China” proposed a deep learning-based framework that incorporates both spatial and temporal features for rainfall forecasting. In this approach, the spatial features are extracted using a Convolutional Neural Network (CNN), while the temporal features are extracted using a Long Short-Term Memory (LSTM) network. This paper presents a deep learning-based spatial-temporal modeling approach for rainfall prediction. The authors used a Convolutional Neural Network (CNN) to extract spatial features from rainfall data, and an LSTM network to capture temporal patterns. They evaluated their model on a large-scale dataset from the Beijing-Tianjin-Hebei region, and demonstrated its superior performance compared to traditional methods.

Another recent study by (Luo et al)\cite{ref13} in the paper “PredRANN: The spatiotemporal attention Convolution Recurrent Neural Network for precipitation nowcasting” proposed a deep learning-based approach for rainfall prediction that incorporates a spatial-temporal attention mechanism. They used a combination of CNN and LSTM networks to extract spatiotemporal features from rainfall data, and then applied an attention mechanism to weight these features. Their experiments on real-world rainfall datasets showed that the proposed approach outperformed existing deep learning models for rainfall prediction.


Deep learning and data-driven methods show promise in enhancing rainfall predictions by incorporating spatiotemporal features and location information. They outperform traditional approaches and have the potential to mitigate risks associated with extreme weather events. This paper explores DL and DMD forecasting, tuning parameters to enhance accuracy.

\section{Methodology}

\subsection{Dataset}

\begin{figure}
\centering
\includegraphics[scale=0.27]{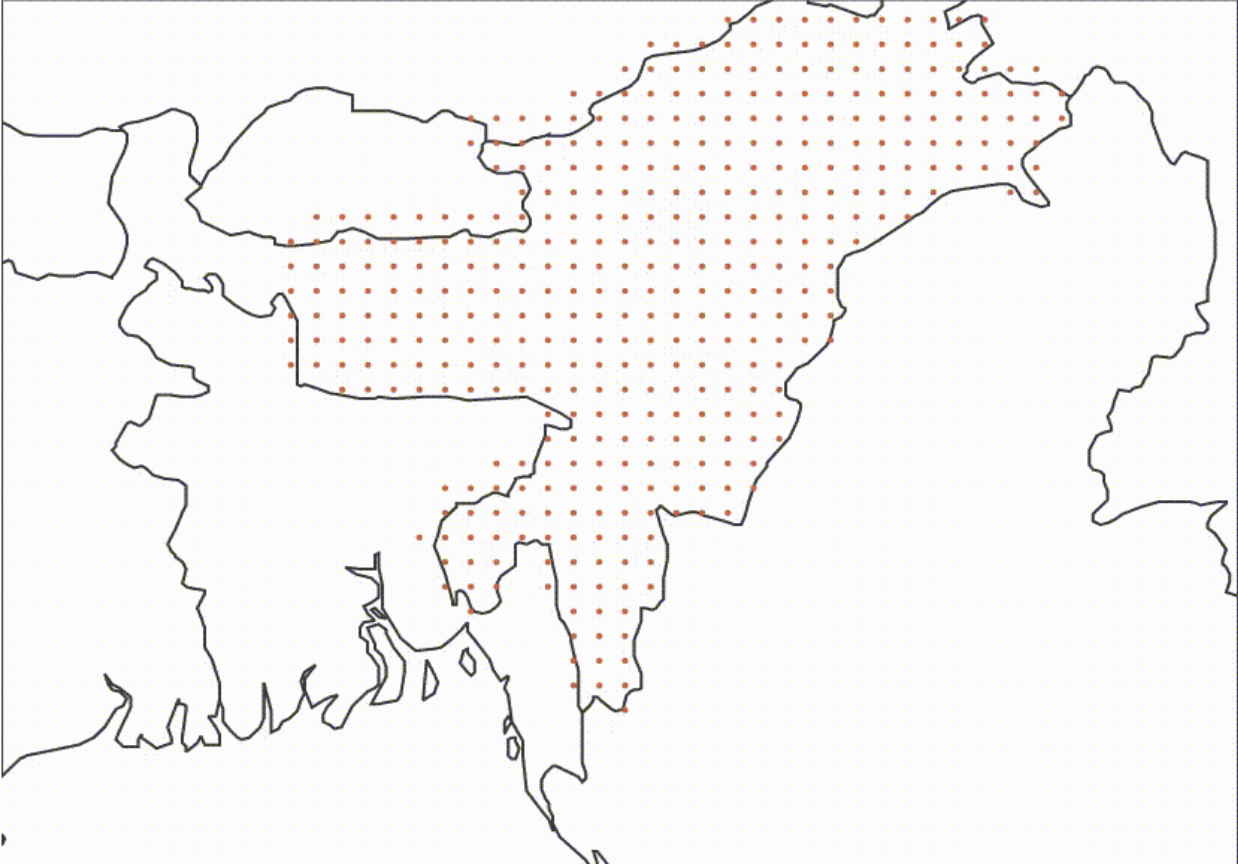}
\caption{Selected Grid Points in North-East India. } \label{gridpoints}
\end{figure}



The dataset used in this study was collected from the India Meteorological Department (IMD) [https://imdpune.gov.in/lrfindex.php] \cite{ref18}. It contains gridded rainfall data with a spatial resolution of 0.25 x 0.25 degree across India for 122 years, from 1901 to 2022. We focused on the North-east region of India from 1901 to 2018, considering 429 grid points between longitudes $89.81^{0}$E to $98^{0}$E and latitudes $21.89^{0}$N to $30^{0}$N (Fig \ref{gridpoints}).

The daily rainfall data is averaged per month to create a new dataset of monthly average rainfall for each of the 429 grid points, used in DMD analysis. For DL techniques, we selected four key locations: Agartala, Guwahati, Imphal, and Itanagar, using monthly average rainfall data from the corresponding four grid points.


\subsection{Data Driven Modelling}

Data-driven modeling uses data to develop accurate predictions and insights in various fields, like engineering and computer science. Unlike traditional methods, it doesn't rely solely on theoretical assumptions, making it flexible and applicable to a wide range of problems. Data-driven models handle large datasets efficiently, enabling researchers to understand complex systems and make precise predictions. In this study, we explore two data-driven techniques, DMD and DL, for time series rainfall forecasting.

\subsubsection{Dynamic Mode Decomposition (DMD)}

Dynamic mode decomposition (DMD) is a data-driven method used to extract the underlying dynamic structures and patterns from complex, high-dimensional systems. DMD works by decomposing the data into a series of modes, each of which represents a spatial-temporal pattern of motion. These modes are determined by the eigenvectors of a matrix constructed from the data, and the associated eigenvalues represent the temporal dynamics of each mode (Tu et al)\cite{ref1}. DMD has been successfully applied in a variety of fields,  in \cite{ref19} the authors have used DMD to identify rice leaf disease and in \cite{ref20} the authors have used DMD to both detect and also classify the defects in Cantilever beams. Overall, DMD is rising in popularity across all the domains and has emerged as a powerful tool for understanding complex, high-dimensional systems and has the potential to revolutionize our understanding of a wide range of phenomena.


The steps for performing DMD are as follows:
\begin{enumerate}

\item 
Collect data from the system you want to analyze. This data should consist of time series measurements of the state variables of the system.

\item
Create a matrix of ``snapshots'' from the data. Each column of the matrix represents a single snapshot of the system's state at a particular point in time. Let the matrix be denoted by $\mathbf{X} = [\mathbf{x}_1, \mathbf{x}_2, \ldots, \mathbf{x}_N]$, where $\mathbf{x}_i \in \mathbb{C}^n$.

\item
Perform Singular Value Decomposition on the matrix of snapshots to obtain its left singular vectors, right singular vectors, and singular values. These will be used to construct the Dynamic Mode Decomposition. \\ Let the SVD of $\mathbf{X}$ be given by $\mathbf{X} = \mathbf{U}\boldsymbol{\Sigma}\mathbf{V}^*$, where $\mathbf{U}$ is a matrix of left singular vectors, $\boldsymbol{\Sigma}$ is a diagonal matrix of singular values, and $\mathbf{V}$ is a matrix of right singular vectors.

\item 
Using the SVD results, construct the DMD modes. These modes are the building blocks of the system's dynamics and describe how the system evolves over time. The DMD modes are given by $\boldsymbol{\Phi} = \mathbf{X}'\mathbf{V}\boldsymbol{\Sigma}^{-1}\mathbf{U}^*$, where $\mathbf{X}'$ is the matrix of snapshots with the last snapshot removed.

\item 
Calculate the eigenvalues of the DMD modes. These eigenvalues represent the frequencies and growth rates of the system's dynamics. The eigenvalues are given by the diagonal elements of the matrix $\boldsymbol{\Lambda} = \mathbf{V}\boldsymbol{\Sigma}^{-1}\mathbf{U}^*\mathbf{X}'\mathbf{X}$.

\item 
Using the DMD modes and eigenvalues, reconstruct the system's dynamics. The time evolution of the system can be approximated as $\mathbf{x}(t) \approx \sum_{k=1}^r \boldsymbol{\phi}_k e^{\omega_k t} b_k$, where $\boldsymbol{\phi}_k$ is the $k$-th DMD mode, $\omega_k$ is the $k$-th eigenvalue, $r$ is the number of significant modes, and $b_k$ is a set of coefficients that can be calculated from the initial conditions of the system.

\end{enumerate}

\begin{figure}
\centering
\includegraphics[scale=0.45]{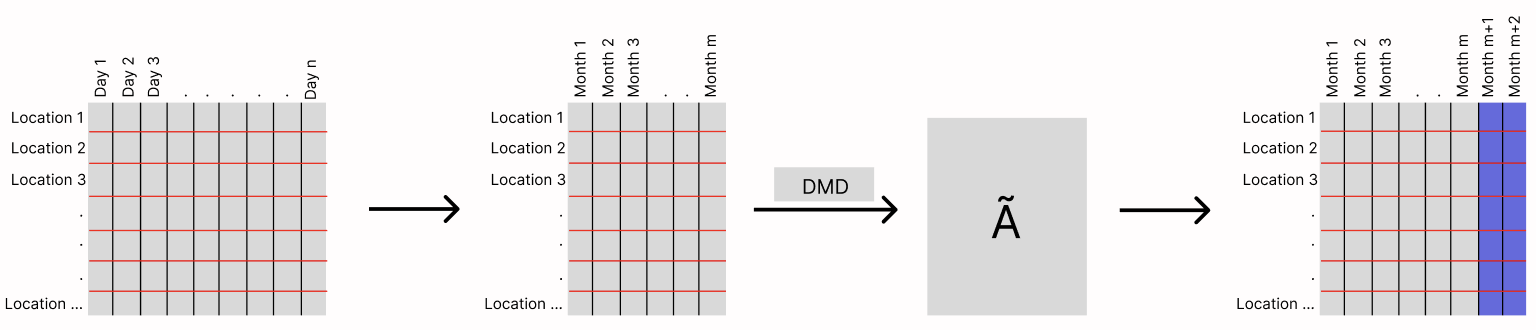}
\caption{Applying DMD to Rainfall data} \label{fig:DMD_workflow}
\end{figure}

Overall, Dynamic Mode Decomposition is a powerful tool for analyzing and understanding complex dynamical systems. It has applications in many fields, including physics, engineering, biology, and finance. Our workflow for DMD analysis is presented in Fig \ref{fig:DMD_workflow}.


\subsection{Deep Learning}

Deep learning is a subset of machine learning that uses artificial neural networks to learn from vast datasets. Its key advantages are automatic feature extraction and handling complex data like images, speech, and language without manual engineering. It excels in tasks like image recognition, speech processing, NLP, and autonomous driving.

Recurrent Neural Networks are a type of neural network designed to process sequential data by retaining and utilizing information from previous time steps. RNNs have been applied in a wide range of applications, including speech recognition, natural language processing, and time series forecasting. 
For example, in \cite{ref21}, authors have used RNN based LSTM model to predict COVID-19 in different states in India And in \cite{ref21}, authors have used RNN based models and sliding window based CNN model to predict the stock price.

One of the main disadvantages of RNNs is the vanishing gradient problem, which occurs when the gradients used to update the weights in the network become too small to have a significant impact on the network's performance. As a result, RNNs may have difficulty learning long-term dependencies and may perform poorly on tasks that require such dependencies. Additionally, RNNs may suffer from overfitting, where the network becomes too complex and begins to memorize the training data instead of generalizing to new data.

Researchers have developed various techniques to mitigate the vanishing gradient problem, including gating mechanisms such as LSTM (Hochreiter \& Schmidhuber)\cite{ref8} and GRU (Chung et al)\cite{ref9} networks, as well as methods for gradient clipping and weight regularization (Pascanu et al)\cite{ref10}.




\begin{figure}
\centering
\includegraphics[scale=0.4]{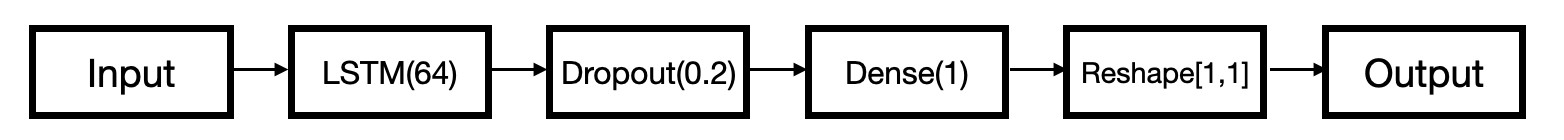}
\caption{Model Architecture} \label{fig:LSTM_arc}
\end{figure}

Long Short Term Memory (LSTM) is a type of recurrent neural network designed to handle the vanishing gradient problem that occurs in traditional RNNs when trying to propagate information over many time steps. LSTM networks consist of a set of memory cells that can selectively store or erase information over time, as well as input, forget, and output gates that control the flow of information into and out of the memory cells. 


The proposed model architecture as shown in Fig \ref{fig:LSTM_arc} consists of the following key components:
LSTM layer with 64 units, capturing long-term dependencies in sequential data.
Dropout layer with a rate of 0.2, reducing overfitting.
Dense layer with 1 unit, initialized with zeros.
Reshape layer to convert the output shape to [1, 1].

This architecture processes sequential data, applies regularization, and outputs a reshaped result.

The input sequence is passed through the LSTM layer, followed by the Dropout layer, Dense layer, and Reshape layer, ultimately resulting in the output.


\begin{figure}
\centering
\includegraphics[scale=0.5]{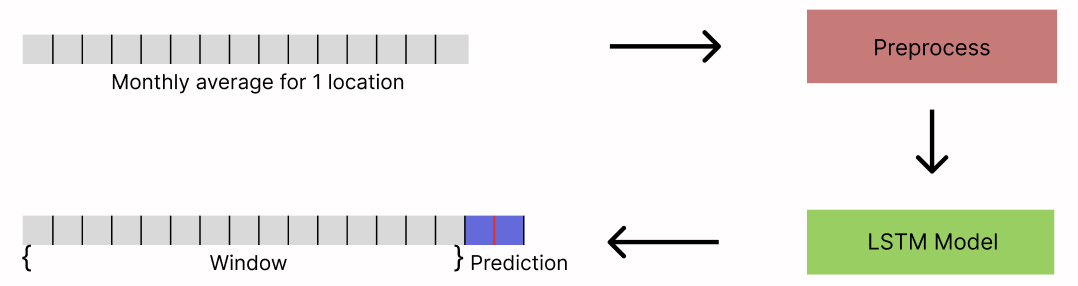} 
\caption{Deep Learning approach to Rainfall Data} \label{fig:LSTM_workflow}
\end{figure}

Fig \ref{fig:LSTM_workflow}, shows the overall workflow of the LSTM approach, The monthly average data is first normalized and is split into windows which is then used to train the LSTM network. After training the network, given a past history window, we will be able to predict rainfall for a future month. 


\subsection{Model Training}

\subsubsection{DMD}

For the dynamic mode decomposition analysis, all locations in the dataset were used, and a subset of locations was also analyzed to test the sensitivity of the results to the number of locations included in the analysis. We also take the monthly average measurements corresponding to each location.
The rows in the Data matrix $X$ will correspond to each location and the columns will correspond to timestamp. In our case it will be for each month. Using this data, we will try to predict for a period of 1 year using a period of 10 years of data for different intervals.

\subsubsection{Deep Learning}

Four locations, Agartala, Imphal, Guwahati, and Itanagar, were selected for deep learning analysis in the northeast region of India. The dataset covered a wide area, spanning from 89.8°E to 98°E and 30°N to 21.89°N. Each location's latitude and longitude values were used to identify them in the dataset. We conducted separate model training for each location using an 80-20 data split. Experiments with various optimizers and dropout layers were performed to improve model accuracy.

\subsubsection{Evaluation metrics}

    RMSE is a common performance metric for regression models that measures the average difference between predicted and actual values. A lower RMSE indicates better predictive performance.

    The formula for the root mean squared error (RMSE) is:

$$\text{RMSE} = \sqrt{\frac{1}{n} \sum_{i=1}^{n} (y_i - \hat{y_i})^2}$$

Here, $n$ represents the number of data points, $y_i$ represents the actual value of the $i$-th data point, and $\hat{y_i}$ represents the predicted value of the $i$-th data point. 

    MAE calculates the average absolute difference between actual and predicted values. It is suitable for data with high variability or outliers, but other metrics and qualitative factors may be considered for comprehensive evaluation.

    The formula for the mean absolute error (MAE) can be written as:
    $$\text{MAE} = \frac{1}{n} \sum_{i=1}^{n} |y_i - \hat{y_i}|$$

    Here, $n$ represents the number of data points, $y_i$ represents the actual value of the $i$-th data point, and $\hat{y_i}$ represents the predicted value of the $i$-th data point.


\section{Results and Discussion}

This section discusses the results obtained using Deep Learning and Dynamic Mode Decomposition (DMD) techniques to predict rainfall. Our results showed that both methods were able to accurately predict rainfall demonstrating the potential for using advanced computational methods to improve weather forecasting and better prepare for extreme weather events.

\subsection{DMD}

Table \ref{DMD_summary_table} shows results from Dynamic Mode Decomposition (DMD) experiments on a rainfall data matrix, constructed from 10 years of data, with predictions made for a single year. DMD was conducted at various projection ranks to obtain a low-dimensional representation of the data. The RMSE and MAE values indicate reasonably accurate rainfall predictions, ranging from 150.44 mm to 263.34 mm and 91.34 mm to 154.61 mm, respectively. Notably, the best performance was observed for the data from 1995-2005.
 \begin{table}[]
\caption{DMD Results for forecasting 1 year}
\centering
\scalebox{\scalevaltab}{
\begin{tabular}{|l|l|l|r|r|r|}
\hline
\begin{tabular}[c]{@{}l@{}}Start\\ Year\end{tabular} &
  \begin{tabular}[c]{@{}l@{}}Stop\\ Year\end{tabular} &
  \begin{tabular}[c]{@{}l@{}}Predicted \\ Duration\end{tabular} &
  \multicolumn{1}{l|}{Rank} &
  \multicolumn{1}{l|}{RMSE} &
  \multicolumn{1}{l|}{MAE} \\ \hline
1929 & 1939 & 1 Year & {  106}                         & {   263.3423} & {   154.6123} \\ \hline
1941 & 1951 & 1 Year & {  123}                         & {   260.2758} & {   144.6304} \\ \hline
1954 & 1964 & 1 Year & {  127}                         & {   177.3236} & {   107.8259} \\ \hline
1973 & 1983 & 1 Year & {  128}                         & {   170.6351} & {   109.3051} \\ \hline
1995 & 2005 & 1 Year &  118 & {   150.4379} & {   91.3362}  \\ \hline
2000 & 2010 & 1 Year & {  100}                         & {   236.3857} & {   124.9998} \\ \hline
2005 & 2015 & 1 Year & {  123}                         & {   158.5830} & {   97.9195}  \\ \hline
\end{tabular}}
\label{DMD_summary_table}

\end{table}



\begin{figure}[h]
\centering



\includegraphics[scale=0.5]{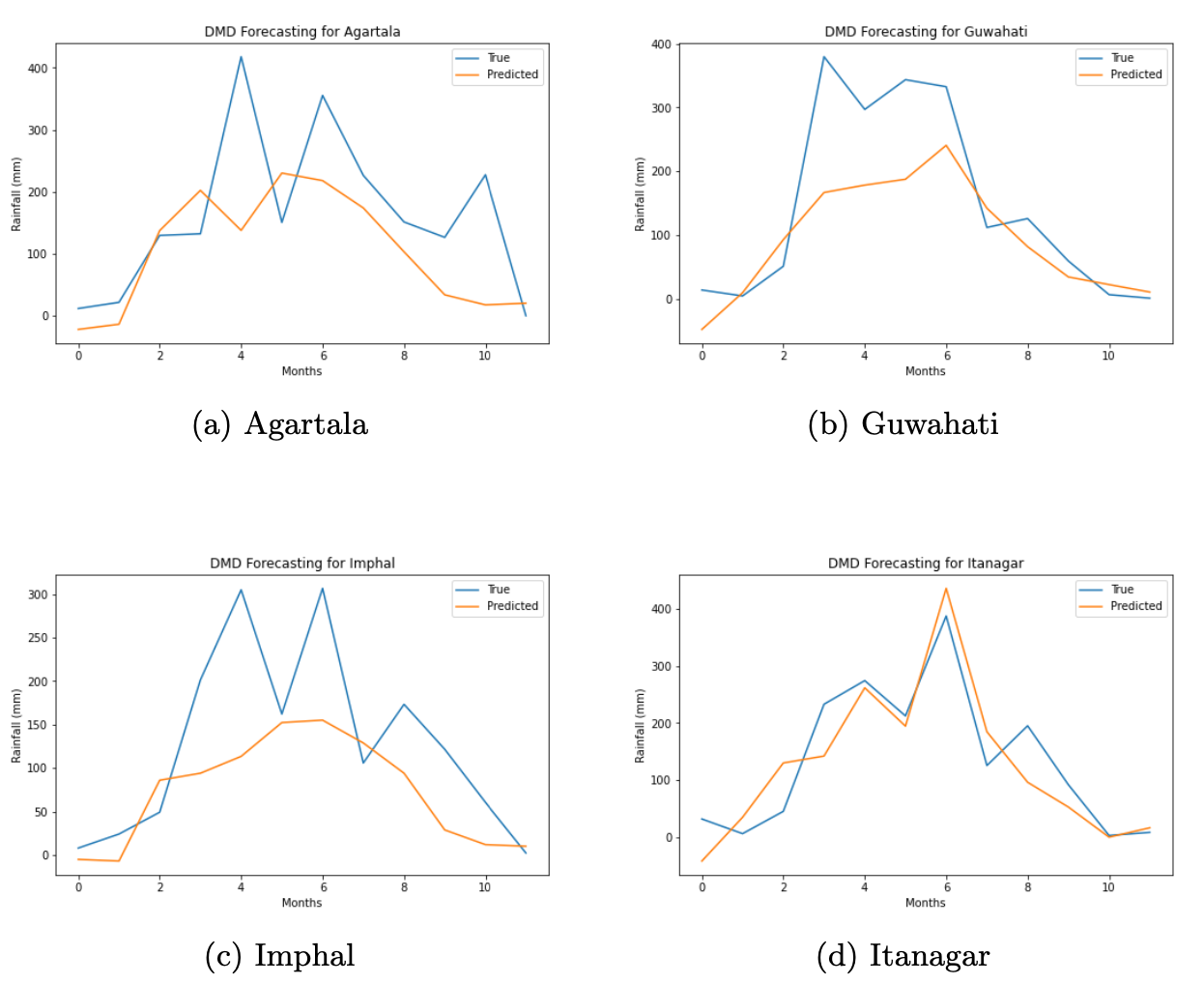}
\caption{Predictions for regions of North East India in 2016 using DMD}
\label{fig:dmd_predictions}
\end{figure}

The rainfall prediction results for the year 2016, for Agartala, Guwahati, Imphal, and Itanagar, were obtained by constructing a matrix $\mathcal{X}$ from rainfall data between 2005 and 2015, using a projection rank of 123. The forecasted precipitation for the entire year of 2016 is shown in Figure \ref{fig:dmd_predictions}. The prediction performance was evaluated using the Mean Squared Error (MSE) and Root Mean Square Error (RMSE), which were calculated to be 97.9195 and 158.5830, respectively.

\subsection{Deep Learning}





This section presents results from the TensorFlow time series prediction model for rainfall in four Indian cities: Itanagar, Guwahati, Agartala, and Imphal. The dataset is split into training and testing sets, and a window generator function is defined to create input and output windows for the model. The model's core includes a single layer of LSTM cells with 64 hidden units, followed by a dropout layer (rate=0.2) to prevent overfitting. Experiments with different optimizers, input/output window sizes, and dropout values were conducted to identify the best-performing model. Evaluation was done using RMSE and MAE metrics. Table \ref{tab:model_params} presents various model parameters.

\begin{table}[h]
\caption{Model Parameters}
\centering
\scalebox{\scalevaltab}{
\begin{tabular}{|c|c|c|}
\hline
Learning Rate & Hidden Units & Loss Function \\
\hline
0.01 & 64 & MAE \\
\hline
\end{tabular}}
\label{tab:model_params}
\end{table}

\begin{table}[!h]
\caption{Performance Analysis of the proposed method for Itanagar}
\centering
\scalebox{\scalevaltab}{
\begin{tabular}{|l|l|l|l|ll|}
\hline
                        &                                             &                     &     & \multicolumn{2}{l|}{ \textbf{Metrics}}   \\ \cline{5-6} 
\multirow{-2}{*}{ \textbf{Optimiser}} &
  \multirow{-2}{*}{\begin{tabular}[c]{@{}l@{}} \textbf{Input}\\  \textbf{Window}\end{tabular}} &
  \multirow{-2}{*}{\begin{tabular}[c]{@{}l@{}} \textbf{Output}\\  \textbf{Window}\end{tabular}} &
  \multirow{-2}{*}{ \textbf{Dropout}} &
  \multicolumn{1}{l|}{ \textbf{MAE}} &
   \textbf{RMSE} \\ \hline
                        &                                             &                     & 0   & \multicolumn{1}{l|}{0.3637} & 0.5285 \\ \cline{4-6} 
                        & \multirow{-2}{*}{13}                        & \multirow{-2}{*}{1} & 0.2 & \multicolumn{1}{l|}{0.3570} & 0.5249 \\ \cline{2-6} 
                        &                                             &                     & 0   & \multicolumn{1}{l|}{0.3650}    & 0.5356    \\ \cline{4-6} 
                        & \multirow{-2}{*}{14}                        & \multirow{-2}{*}{1} & 0.2 & \multicolumn{1}{l|}{0.3612}    & 0.4527    \\ \cline{2-6} 
                        &                                             &                     & 0   & \multicolumn{1}{l|}{0.3792}    & 0.5505    \\ \cline{4-6} 
\multirow{-6}{*}{AdamW} & \multirow{-2}{*}{15}                        & \multirow{-2}{*}{1} & 0.2 & \multicolumn{1}{l|}{0.3812}    & 0.5459    \\ \hline
                        & {  }                     &                     & \textbf{0}   & \multicolumn{1}{l|}{\textbf{0.3526}}    & \textbf{0.5260}   \\ \cline{4-6} 
                        & \multirow{-2}{*}{{  \textbf{13}}} & \multirow{-2}{*}{\textbf{1}} & 0.2 & \multicolumn{1}{l|}{0.3550}    & 0.5258    \\ \cline{2-6} 
                        & {  }                     &                     & 0   & \multicolumn{1}{l|}{0.3569}    & 0.5344    \\ \cline{4-6} 
                        & \multirow{-2}{*}{{  14}} & \multirow{-2}{*}{1} & 0.2 & \multicolumn{1}{l|}{0.3695}    & 0.5372    \\ \cline{2-6} 
                        & {  }                     &                     & 0   & \multicolumn{1}{l|}{0.3670}    &  0.5412   \\ \cline{4-6} 
\multirow{-6}{*}{\textbf{Nadam}} &
  \multirow{-2}{*}{{ 15}} &
  \multirow{-2}{*}{1} &
  0.2 &
  \multicolumn{1}{l|}{0.3707} & 0.5467
   \\ \hline
\end{tabular}}
\label{tab:itanagarDl}
\end{table}

\begin{figure}
\includegraphics[width=\linewidth]{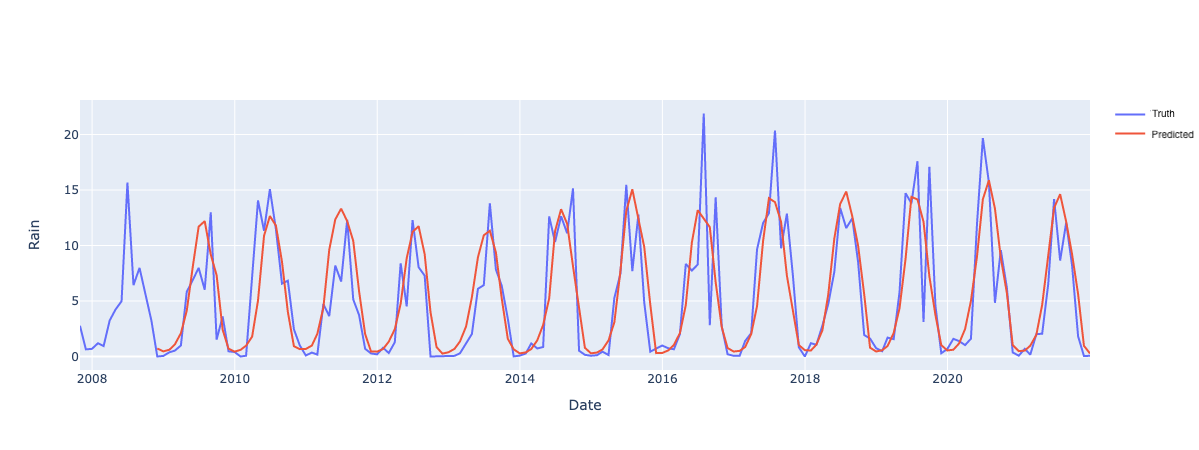}
\caption{Prediction Visualisation for Itanagar} \label{itanagar_model_pred}
\end{figure}

Table \ref{tab:itanagarDl} shows the results of experiments for Itanagar. The best performing optimizer and parameter combination for MAE is Nadam with input window size of 13, output window size of 1, and dropout of 0.2 with a value of 0.3707, and for RMSE is AdamW with input window size of 14, output window size of 1, and dropout of 0.2 with a value of 0.4527. The figure \ref{itanagar_model_pred} shows the performance of the model while predicting on the test set. 

\begin{table}[!h]
\caption{Performance Analysis of the proposed method for Imphal}
\centering
\scalebox{\scalevaltab}{
\begin{tabular}{|l|l|l|l|ll|}
\hline
                        &                                             &                     &     & \multicolumn{2}{l|}{ \textbf{Metrics}}   \\ \cline{5-6} 
\multirow{-2}{*}{ \textbf{Optimiser}} &
  \multirow{-2}{*}{\begin{tabular}[c]{@{}l@{}} \textbf{Input}\\  \textbf{Window}\end{tabular}} &
  \multirow{-2}{*}{\begin{tabular}[c]{@{}l@{}} \textbf{Output}\\  \textbf{Window}\end{tabular}} &
  \multirow{-2}{*}{ \textbf{Dropout}} &
  \multicolumn{1}{l|}{ \textbf{MAE}} &
   \textbf{RMSE} \\ \hline
                        &                                             &                     & 0   & \multicolumn{1}{l|}{0.4422} & 0.6505 \\ \cline{4-6} 
                        & \multirow{-2}{*}{13}                        & \multirow{-2}{*}{1} & 0.2 & \multicolumn{1}{l|}{0.4421} & 0.6508 \\ \cline{2-6} 
                        &                                             &                     & 0   & \multicolumn{1}{l|}{0.4448}    & 0.6533    \\ \cline{4-6} 
                        & \multirow{-2}{*}{14}                        & \multirow{-2}{*}{1} & 0.2 & \multicolumn{1}{l|}{0.4477}    & 0.6561    \\ \cline{2-6} 
                        &                                             &                     & 0   & \multicolumn{1}{l|}{0.4513}    & 0.6593    \\ \cline{4-6} 
\multirow{-6}{*}{AdamW} & \multirow{-2}{*}{15}                        & \multirow{-2}{*}{1} & 0.2 & \multicolumn{1}{l|}{0.4519}    & 0.6583    \\ \hline
                        & {  }                     &                     & 0   & \multicolumn{1}{l|}{0.4434}    &  0.6534    \\ \cline{4-6} 
                        & \multirow{-2}{*}{{  13}} & \multirow{-2}{*}{1} & 0.2 & \multicolumn{1}{l|}{0.4378}    & 0.6449    \\ \cline{2-6} 
                        & {  }                     &                     & 0   & \multicolumn{1}{l|}{0.4424}    & 0.6490    \\ \cline{4-6} 
                        & \multirow{-2}{*}{{  \textbf{14}}} & \multirow{-2}{*}{\textbf{1}} & \textbf{0.2} & \multicolumn{1}{l|}{\textbf{0.4357}}    & \textbf{0.6419}    \\ \cline{2-6} 
                        & {  }                     &                     & 0   & \multicolumn{1}{l|}{0.4381}    &  0.6446   \\ \cline{4-6} 
\multirow{-6}{*}{\textbf{Nadam}} &
  \multirow{-2}{*}{{ 15}} &
  \multirow{-2}{*}{1} &
  0.2 &
  \multicolumn{1}{l|}{0.4423} & 0.6502
   \\ \hline
\end{tabular}}
\label{imphal_dl_pred}
\end{table}

\begin{figure}[!h]
\includegraphics[width=\textwidth]{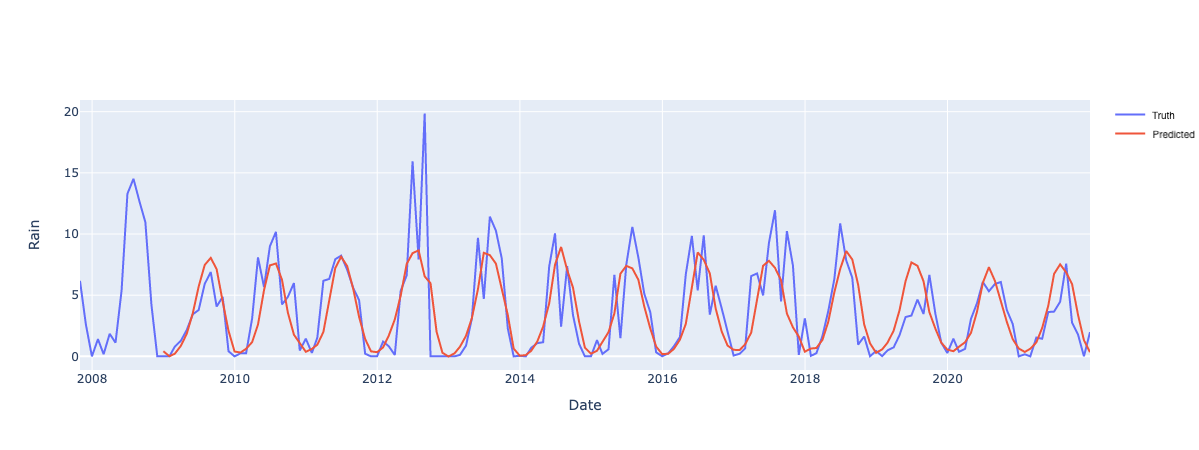}
\caption{Prediction Visualisation for Imphal} \label{imphal_model_pred}
\end{figure}

Table \ref{tab:itanagarDl} shows the results of experiments for Imphal.The results suggest that the combination of Nadam optimizer, an input window size of 14, an output window size of 1, and a dropout rate of 0.2 achieved the lowest MAE and RMSE values. The figure \ref{imphal_model_pred} shows the performance of the model while predicting on the test set.

\begin{table}[!h]
\caption{Performance Analysis of the proposed method for Guhawati}
\centering
\scalebox{\scalevaltab}{
\begin{tabular}{|l|l|l|l|ll|}
\hline
                        &                                             &                     &     & \multicolumn{2}{l|}{ \textbf{Metrics}}   \\ \cline{5-6} 
\multirow{-2}{*}{ \textbf{Optimiser}} &
  \multirow{-2}{*}{\begin{tabular}[c]{@{}l@{}} \textbf{Input}\\  \textbf{Window}\end{tabular}} &
  \multirow{-2}{*}{\begin{tabular}[c]{@{}l@{}} \textbf{Output}\\  \textbf{Window}\end{tabular}} &
  \multirow{-2}{*}{ \textbf{Dropout}} &
  \multicolumn{1}{l|}{ \textbf{MAE}} &
   \textbf{RMSE} \\ \hline
                        &                                             &                     & 0   & \multicolumn{1}{l|}{0.3402} & 0.4763 \\ \cline{4-6} 
                        & \multirow{-2}{*}{13}                        & \multirow{-2}{*}{1} & 0.2 & \multicolumn{1}{l|}{0.3368} & 0.4827 \\ \cline{2-6} 
                        &                                             &                     & 0   & \multicolumn{1}{l|}{0.3341}    & 0.4707    \\ \cline{4-6} 
                        & \multirow{-2}{*}{14}                        & \multirow{-2}{*}{1} & 0.2 & \multicolumn{1}{l|}{0.3374}    & 0.4668    \\ \cline{2-6} 
                        &                                             &                     & 0   & \multicolumn{1}{l|}{0.3351}    & 0.4746    \\ \cline{4-6} 
\multirow{-6}{*}{AdamW} & \multirow{-2}{*}{15}                        & \multirow{-2}{*}{1} & 0.2 & \multicolumn{1}{l|}{0.3371}    & 0.4757    \\ \hline
                        & {  }                     &                     & 0   & \multicolumn{1}{l|}{0.3264}    & 0.4665    \\ \cline{4-6} 
                        & \multirow{-2}{*}{{  \textbf{13}}} & \multirow{-2}{*}{\textbf{1}} & \textbf{0.2} & \multicolumn{1}{l|}{\textbf{0.2990}}    & \textbf{0.4410}    \\ \cline{2-6} 
                        & {  }                     &                     & 0   & \multicolumn{1}{l|}{0.3115}    & 0.4555    \\ \cline{4-6} 
                        & \multirow{-2}{*}{{  14}} & \multirow{-2}{*}{1} & 0.2 & \multicolumn{1}{l|}{0.3091}    & 0.4559    \\ \cline{2-6} 
                        & {  }                     &                     & 0   & \multicolumn{1}{l|}{0.3126}    &  0.4695   \\ \cline{4-6} 
\multirow{-6}{*}{\textbf{Nadam}} &
  \multirow{-2}{*}{{ 15}} &
  \multirow{-2}{*}{1} &
  0.2 &
  \multicolumn{1}{l|}{0.3126} & 0.4525
   \\ \hline
\end{tabular}}
\label{tab:guwahatiDL}
\end{table}

\begin{figure}[!h]
\includegraphics[width=\textwidth]{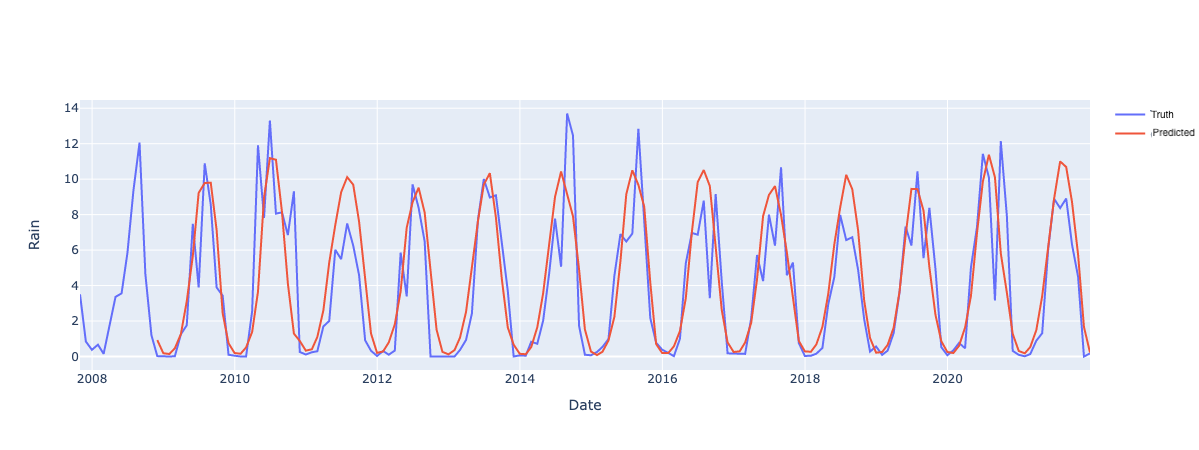}
\caption{Prediction Visualisation for Guwahati} \label{guwahati_model_pred}
\end{figure}

Table \ref{tab:guwahatiDL} shows the results of experiments for Guwahati.The results suggest that the combination of Nadam optimizer, an input window size of 13, an output window size of 1, and a dropout rate of 0.2 achieved the lowest MAE and RMSE values. The figure \ref{guwahati_model_pred} shows the performance of the model while predicting on the test set.

\begin{table}[]
\caption{Performance Analysis of the proposed method for Agartala}
\centering
\scalebox{\scalevaltab}{
\begin{tabular}{|l|l|l|l|ll|}
\hline
                        &                                             &                     &     & \multicolumn{2}{l|}{ \textbf{Metrics}}   \\ \cline{5-6} 
\multirow{-2}{*}{ \textbf{Optimiser}} &
  \multirow{-2}{*}{\begin{tabular}[c]{@{}l@{}} \textbf{Input}\\  \textbf{Window}\end{tabular}} &
  \multirow{-2}{*}{\begin{tabular}[c]{@{}l@{}} \textbf{Output}\\  \textbf{Window}\end{tabular}} &
  \multirow{-2}{*}{ \textbf{Dropout}} &
  \multicolumn{1}{l|}{ \textbf{MAE}} &
   \textbf{RMSE} \\ \hline
                        &                                             &                     & 0   & \multicolumn{1}{l|}{0.3284} & 0.5322 \\ \cline{4-6} 
                        & \multirow{-2}{*}{13}                        & \multirow{-2}{*}{1} & 0.2 & \multicolumn{1}{l|}{0.3887} & 0.5436 \\ \cline{2-6} 
                        &                                             &                     & 0   & \multicolumn{1}{l|}{0.3859}    & 0.5394    \\ \cline{4-6} 
                        & \multirow{-2}{*}{14}                        & \multirow{-2}{*}{1} & 0.2 & \multicolumn{1}{l|}{0.3839}    & 0.5388    \\ \cline{2-6} 
                        &                                             &                     & 0   & \multicolumn{1}{l|}{0.3901}    & 0.5451   \\ \cline{4-6} 
\multirow{-6}{*}{AdamW} & \multirow{-2}{*}{15}                        & \multirow{-2}{*}{1} & 0.2 & \multicolumn{1}{l|}{0.3820}    & 0.5324    \\ \hline
                        & {  }                     &                     & 0   & \multicolumn{1}{l|}{0.3786}    & 0.5314    \\ \cline{4-6} 
                        & \multirow{-2}{*}{{  13}} & \multirow{-2}{*}{1} & 0.2 & \multicolumn{1}{l|}{0.3709}    & 0.5227    \\ \cline{2-6} 
                        & {  }                     &                     & \textbf{0}   & \multicolumn{1}{l|}{\textbf{0.3698}}    & \textbf{0.5271}    \\ \cline{4-6} 
                        & \multirow{-2}{*}{{ \textbf{14}}} & \multirow{-2}{*}{\textbf{1}} & 0.2 & \multicolumn{1}{l|}{0.3770}    & 0.5348    \\ \cline{2-6} 
                        & {  }                     &                     & 0   & \multicolumn{1}{l|}{0.3804}    &  0.5388   \\ \cline{4-6} 
\multirow{-6}{*}{\textbf{Nadam}} &
  \multirow{-2}{*}{{ 15}} &
  \multirow{-2}{*}{1} &
  0.2 &
  \multicolumn{1}{l|}{0.3840} & 0.5396
   \\ \hline
\end{tabular}}
\label{tab:agartalaDL}
\end{table}

\begin{figure}[!h]
\includegraphics[width=\textwidth]{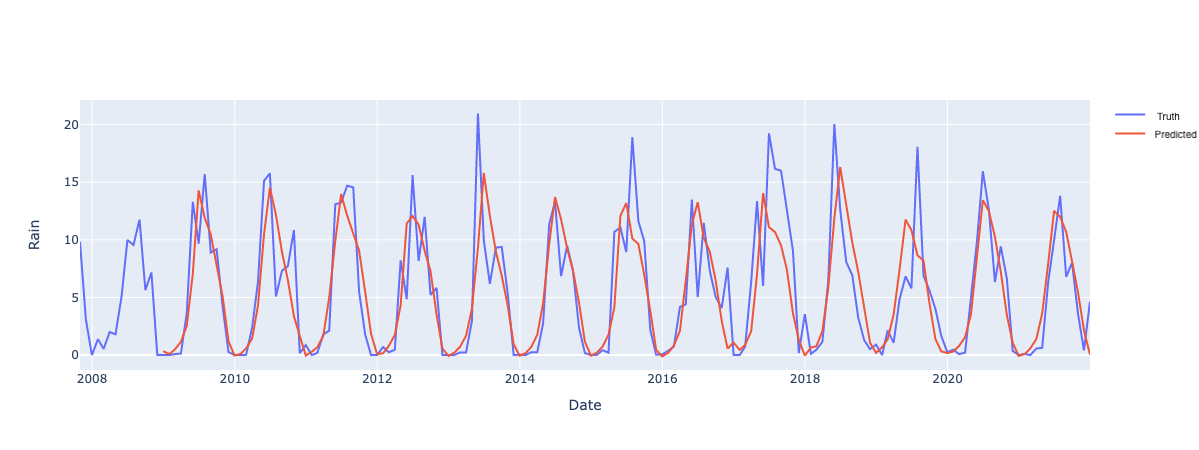}
\caption{Prediction Visualisation for Agartala } \label{agartala_model_pred}
\end{figure}

Table \ref{tab:agartalaDL} shows the results of experiments for Agartala.The results suggest that the combination of Nadam optimizer, an input window size of 14, an output window size of 1, and a dropout rate of 0 achieved the lowest MAE and RMSE values. The figure \ref{agartala_model_pred} shows the performance of the model while predicting on the test set.

\section{Conclusion}
Our findings suggest that both Dynamic Mode Decomposition (DMD) and the proposed LSTM neural network, are capable of accurately predicting rainfall. However, the LSTM model's ability to use memory cells to analyze and capture longer-term patterns and spatiotemporal dependencies allows it to forecast rainfall peaks effectively compared to the DMD model. This enables it to give early warning for potential flood events. Early warning triggers can provide crucial information to decision-makers and emergency responders, allowing them to take timely actions to mitigate the impact of natural disasters.
Overall, the results of this study demonstrate the potential of LSTM to improve rainfall prediction accuracy and to provide early warning triggers for effective disaster preparedness and mitigation, especially in the more vulnerable communities.

\bibliographystyle{splncs04}
\bibliography{refs.bib}

\end{document}